\documentclass[a4paper,conference]{IEEEtran}

\usepackage{array}
\usepackage{epsfig}
\usepackage{graphicx}
\usepackage{amsmath}
\usepackage{amssymb}
\usepackage{multirow}
\usepackage{makecell}
\usepackage[breaklinks=true,bookmarks=false]{hyperref}
\usepackage{authblk}

\begin{document}
%
\title{Efficient grouping for keypoint detection}


\author[ ]{
    Alexey Sidnev\textsuperscript{1,2},
    Ekaterina Krasikova\textsuperscript{1},
	Maxim Kazakov\textsuperscript{1,3}
}
\affil[1]{Huawei Research Center, Nizhny Novgorod, Russia}
\affil[2]{Lobachevsky State University of Nizhny Novgorod, Russia}
\affil[3]{National Research University Higher School of Economics, Nizhny Novgorod, Russia}
\affil[ ]{\texttt{\footnotesize {\{sidnev.alexey, krasikova.ekaterina, kazakov.maxim\}@huawei.com}}}

\maketitle

\begin{abstract}
The success of deep neural networks in the traditional keypoint detection task encourages researchers to solve new problems and collect more complex datasets.
The size of the DeepFashion2 dataset~\cite{DeepFashion2} poses a new challenge on the keypoint detection task,
as it comprises 13 clothing categories that span a wide range of keypoints (294 in total).
The direct prediction of all keypoints leads to huge memory consumption, slow training, and a slow inference time.
This paper studies the keypoint grouping approach and how it affects the performance of the CenterNet~\cite{CenterNet} architecture.
We propose a simple and efficient automatic grouping technique with a powerful
post-processing method and apply it to the DeepFashion2 fashion landmark task and the MS COCO pose estimation task.
This reduces memory consumption and processing time during inference by up to 19\% and 30\% respectively, and during the training stage by 28\% and 26\% respectively, without compromising accuracy.
\end{abstract}


%
\IEEEpeerreviewmaketitle

\section{Introduction}\label{sec:introduction}

Recent research shows that keypoints, which are also known as landmarks, are one of the most distinctive and robust representations of visual fashion analysis.
The class of keypoint-based methods in computer vision includes the detection and further processing of keypoints.
They can be utilized for object detection, pose estimation, facial landmark recognition, and more.

\begin{figure}[t]
	\centering
	\includegraphics[width=\columnwidth]{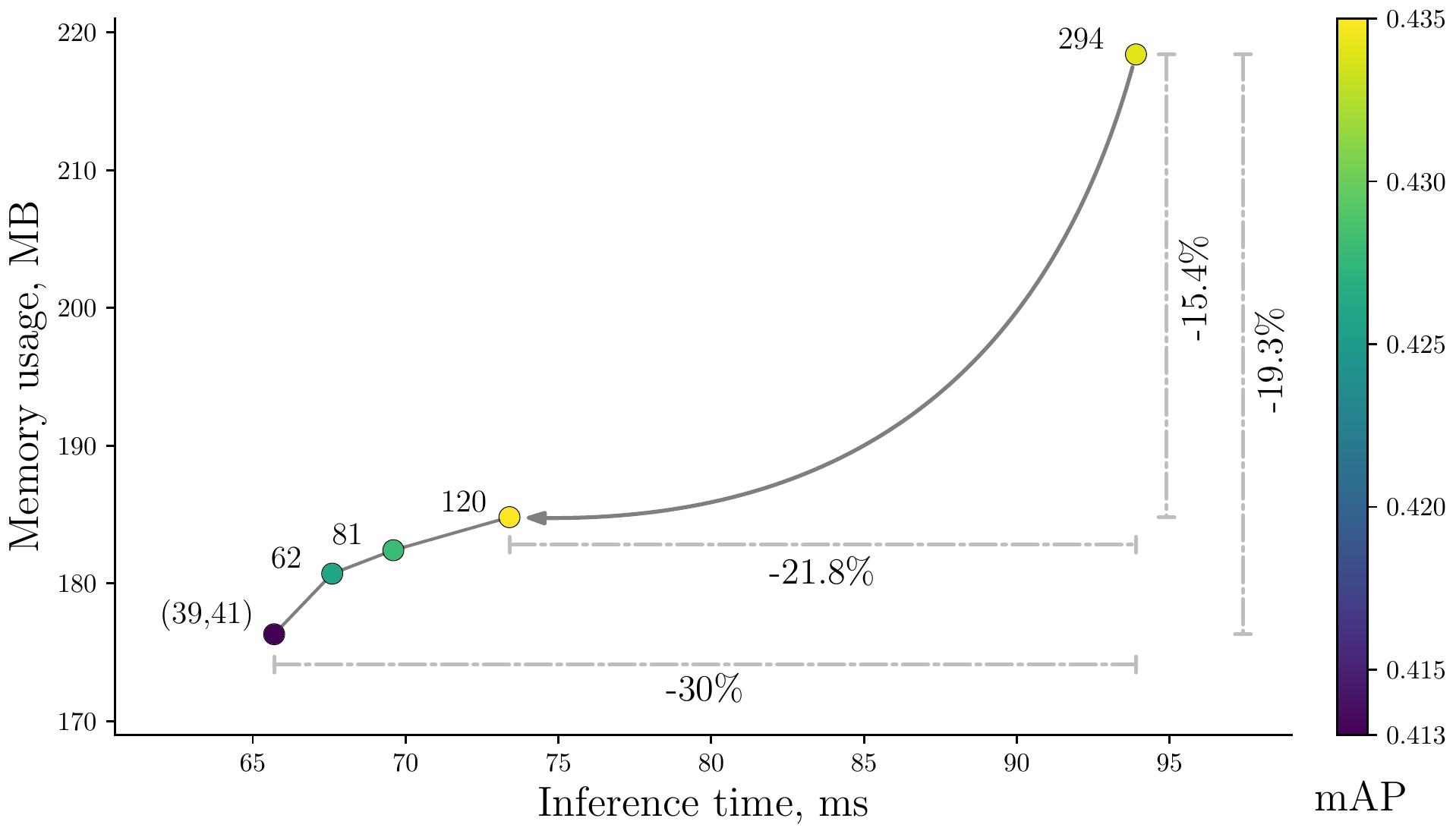}
	\caption{The inference time and memory usage of the mDLA-34 model~\cite{DeepMarkCC} on the DeepFashion2 dataset.
	Performance is shown for models with a direct prediction of 294 keypoints, and
		with groupings obtained through the proposed approach.
	Color scale represents the accuracy on the keypoint detection task.
	Performance is measured on Huawei Mate 20 (Kirin 980)
		with MNN v1.0.1 (number of threads = 4)~\cite{alibaba2020mnn}.}
	\label{fig:time_mem_phone_plot}
\end{figure}

The performance of keypoint detection models strongly depends on the number of unique keypoints defined in a task,
and this could be problematic for large datasets.
One of the newest fashion datasets DeepFashion2 provides annotations for 13 classes, each of which is characterized by a certain set of keypoints, totaling 294.
Therefore, the straightforward prediction of all keypoints requires a heavy CNN architecture, leading to huge memory consumption, slow training, and low inference speed (see Figure~\ref{fig:time_mem_phone_plot}),
which restricts the application areas of such a model.

Authors~\cite{AggregationAF,DeepMarkCC} use semantic information from the task to manually merge certain keypoints into one group.
This helps train the model more quickly while also consuming less memory.
However, the manual grouping may contain human errors and in many cases may not be optimal.

In this paper, we propose and study various techniques of automatic keypoint grouping which help improve the CNN model in terms of training speed, inference time,
and memory consumption without compromising accuracy.
Additionally, we study a powerful post-processing technique specifically designed to improve the accuracy of keypoint detection after grouping.
We show the results of the clothing landmark detection task on the DeepFashion2 dataset~\cite{DeepFashion2} and those of the human pose estimation task on the MS COCO dataset~\cite{lin2014microsoft}.

\section{Related Work}

In general, the keypoint detection task can be used in numerous application scenarios, such as to identify human poses~\cite{alej2016stacked}, find facial landmarks~\cite{zhang2014facial},
and estimate clothing landmarks~\cite{DeepFashion2, liu2016deepfashion, zheng2018modanet}.

DeepFashion2~\cite{DeepFashion2} poses a new challenge on the keypoint detection task.
It is a large-scale fashion dataset that contains consumer-commercial image pairs, and labels such as clothing attributes,
landmarks, and segmentation masks.
The public version of DeepFashion2 train set contains 191,961 rich images covering 13 popular clothing categories from both commercial shopping stores and consumers.
It has about 320 thousand clothing items, and the number of keypoints for every category varies from 8 to 39, with 294 unique keypoints in total.

The keypoint detection task for multiple objects can be solved in a number of ways:
\begin{itemize}
	\item Top-down:
	We first detect objects, and then estimate the keypoint position for each object~\cite{fang2017rmpe, MaskRCNN, CenterNet}.
	\item Bottom-up:
	We first detect all keypoints on the image, and then group the keypoints into objects~\cite{cao2018openpose, pishchulin2016deepcut}.
\end{itemize}

Keypoint-based detection methods are becoming more popular in recent research as they are simpler, faster,
and more accurate compared to anchor-based detectors.
Previous approaches like~\cite{liu2016ssd,redmon2016you} require the manual designing of anchor boxes to train a detector.
The subsequent approach involved a series of anchor-free object detectors, where the goal was to predict the keypoints of the bounding box,
rather than trying to fit an object to an anchor.
Law and Deng proposed a novel anchor-free framework CornerNet~\cite{law2018cornernet}, which detects objects as a pair of corners.
On each position of the feature map, class heatmaps, pair embeddings and corner offsets were predicted.
Class heatmaps calculated the probabilities of pixels being corners of an object, and corner offsets were used to regress the corner locations
while the pair embeddings grouped a pair of corners that belong to the same objects.
Without relying on manually designed anchors to match objects, CornerNet significantly improved detection accuracy on the MS COCO dataset.
Subsequently, there were several other variants of keypoint-based one-stage detectors, with one of them being CenterNet~\cite{CenterNet}.

\begin{table}
	\caption{Number of output channels for the CenterNet model, with 294 directly predicted keypoints for DeepFashion2}
	\label{tab:channels}
	\begin{center}
		\begin{tabular}{|c|c|c|}
			\hline
			Output tensor & Number of channels  & Channel type \\
			\hline
			Center heatmap & $C=13$ & Heatmap \\
			\hline
			Center offset & $ 2 = 1 \cdot 2 $ & Regression: $\bigtriangleup$x, $\bigtriangleup$y \\
			\hline
			Object size & $ 2 = 1 \cdot 2 $ & Regression: w, h \\
			\hline
			Keypoint regression & $ 588 = 294 \cdot 2 $ & Regression: $\bigtriangleup$x, $\bigtriangleup$y \\
			\hline
			Keypoint heatmap & $K=294$ & Heatmap \\
			\hline
			Keypoint offset & $ 2 = 1 \cdot 2 $ & Regression: $\bigtriangleup$x, $\bigtriangleup$y \\
			\hline
		\end{tabular}
	\end{center}
\end{table}

Methods that utilize hierarchical information to improve the accuracy and performance of the model were widely studied in the domain of image
classification~\cite{marszalek2007semantic, deng2014large, zhang2016embedding}.
Some of them explicitly used hierarchical information or encoded the properties of such a class hierarchy into the probabilistic model,
while others attempted to address severe mistakes by using graph distances in class hierarchies~\cite{verma2012learning, deng2010does, zhao2011large}.
Visual hierarchies can be learned or used implicitly.
The same general idea of leveraging external semantic information to improve performance may be applied to the keypoint detection task.

Authors~\cite{AggregationAF,DeepMarkCC} use semantic information in the clothes landmark estimation task of the DeepFashion2 dataset
to achieve grouping by manually merging certain keypoints into one group.
Some clothing landmarks are evidently a subset of others.
For example, shorts can be represented as a part of trousers;
therefore, they do not need unique keypoints and can be merged into trousers.
As such, we can formulate a rule for semantic grouping: the semantically identical keypoints (collar center, top sleeve edge, etc.) from different categories are merged into one group.
This grouping method helps train the model faster, while using less memory.

\section{Keypoint Grouping}

\begin{figure}[t]
	\centering
	\includegraphics[width=\columnwidth]{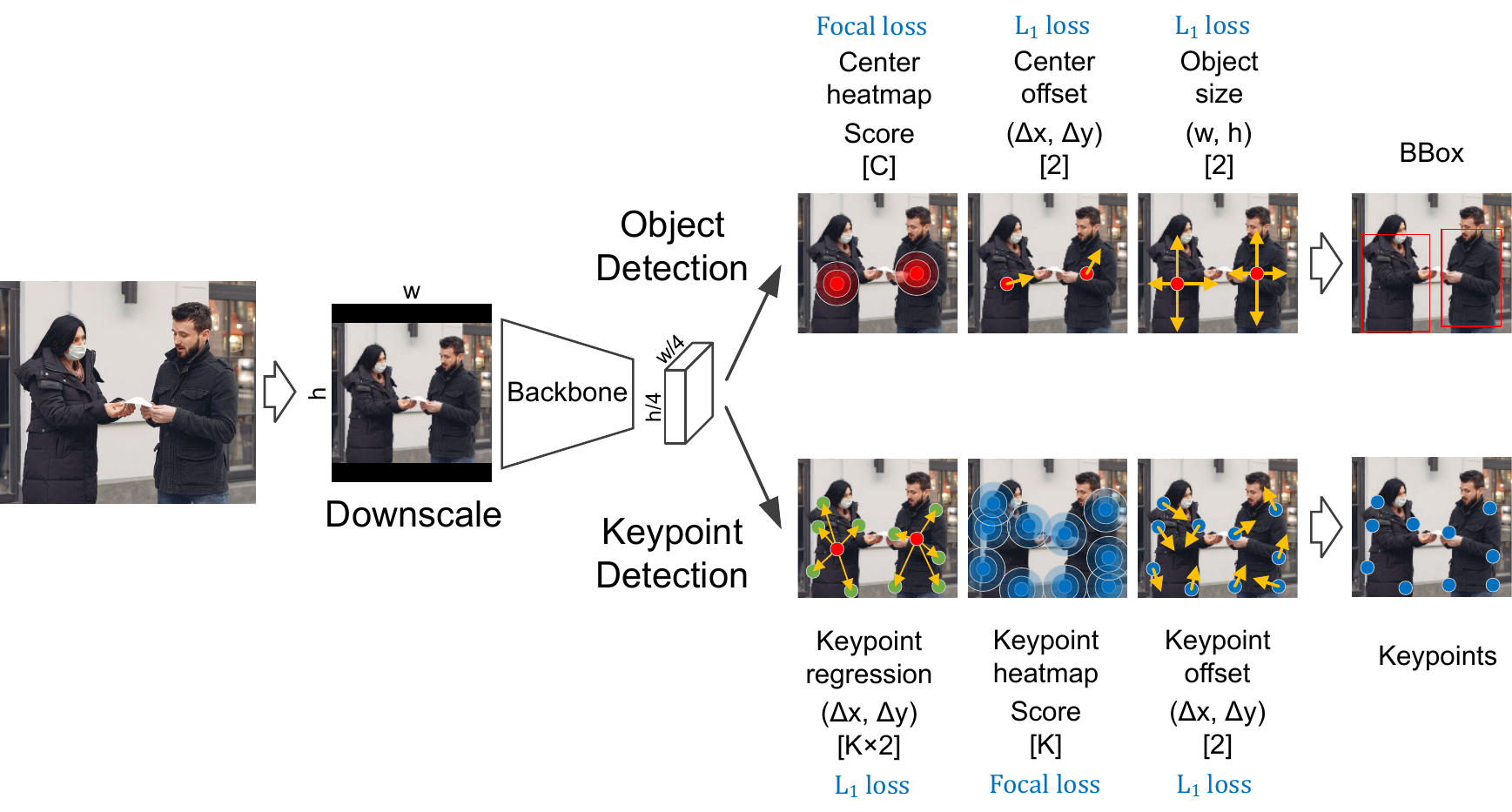}
	\caption{CenterNet architecture for keypoint detection.}
	\label{fig:scheme}
\end{figure}

\subsection{CenterNet Architecture}

CenterNet~\cite{CenterNet} architecture has proven to be effective for a wide range of tasks. In particular, it is able to carry out two tasks simultaneously:
object detection and keypoint location estimation.
Figure~\ref{fig:scheme} illustrates this architecture, where a backbone network downsamples an image four times to generate a feature map,
which is then processed to identify objects and corresponding keypoints.
The proposed architecture has six output tensors, also called heads.

A center heatmap is used to predict the probability of each pixel being the object's center for each of $C$ classes.
The center of an object is defined as the center of a bounding box.
A ground truth heatmap is generated by applying a Gaussian function at each object's center.
Two additional channels in the output feature map $\bigtriangleup$x and $\bigtriangleup$y are used to refine the center coordinates,
while both width and height are predicted directly.

Another branch handles keypoint estimation.
This task involves estimating 2D keypoint locations for each object in one image.
The coarse locations of the keypoints are regressed as relative displacements from the center of the box (keypoint regression in Figure~\ref{fig:scheme}).
Consequently, if a certain pixel has already been classified as an object's center,
you can take the values in the same spatial location from this tensor and interpret them as vectors to keypoints.
Keypoint positions obtained through regression are not entirely accurate, and as such, an additional heatmap with
probabilities is used for each keypoint type to refine the corresponding locations.
Here, a local maximum with high confidence in the heatmap is used as a refined keypoint position.
Like the detection case, two additional channels ($\bigtriangleup$x and $\bigtriangleup$y) are used to obtain more precise keypoint coordinates.
During model inference, the location of each coarse keypoint is replaced with the closest refined keypoint position.
In this way, we can group keypoints belonging to the same object.

During the training stage, 
CenterNet uses focal loss for heatmaps and L1 loss for each regression feature map.
The loss function for keypoint regression is computed only for keypoints that are presented in the ground truth.

\subsection{Grouping Analysis}

One of the first steps to overcoming the challenge in keypoint detection involves defining the model output.
A straightforward approach is to concatenate keypoints from every category and deal with them separately.
However, this is not the best solution because of the huge size of the model output.
For instance, directly predicting 294 keypoints leads to a huge number of output channels: $ 901 = 13 + 2 + 2 + 588 + 294 + 2$.
In this case, two tensors for keypoint detection occupy 97.9 \% of output channels and computations.

To store FP32 output activations for an input resolution of $512 \times 512$, 56 MB is needed ($512 / 4 \cdot 512 / 4 \cdot 901 \cdot 4$).
This contributes to over 20\% of the total activation size for ResNet-50 and DLA-34.
In Table~\ref{tab:activation_size}, we can assume that ReLU and BatchNorm operations do not occupy extra memory, while the last column shows
how much memory is occupied by the output tensors as a percentage of the total memory consumption (activations + weights + input).

\begin{figure}[t]
	\centering
	\includegraphics[width=\columnwidth]{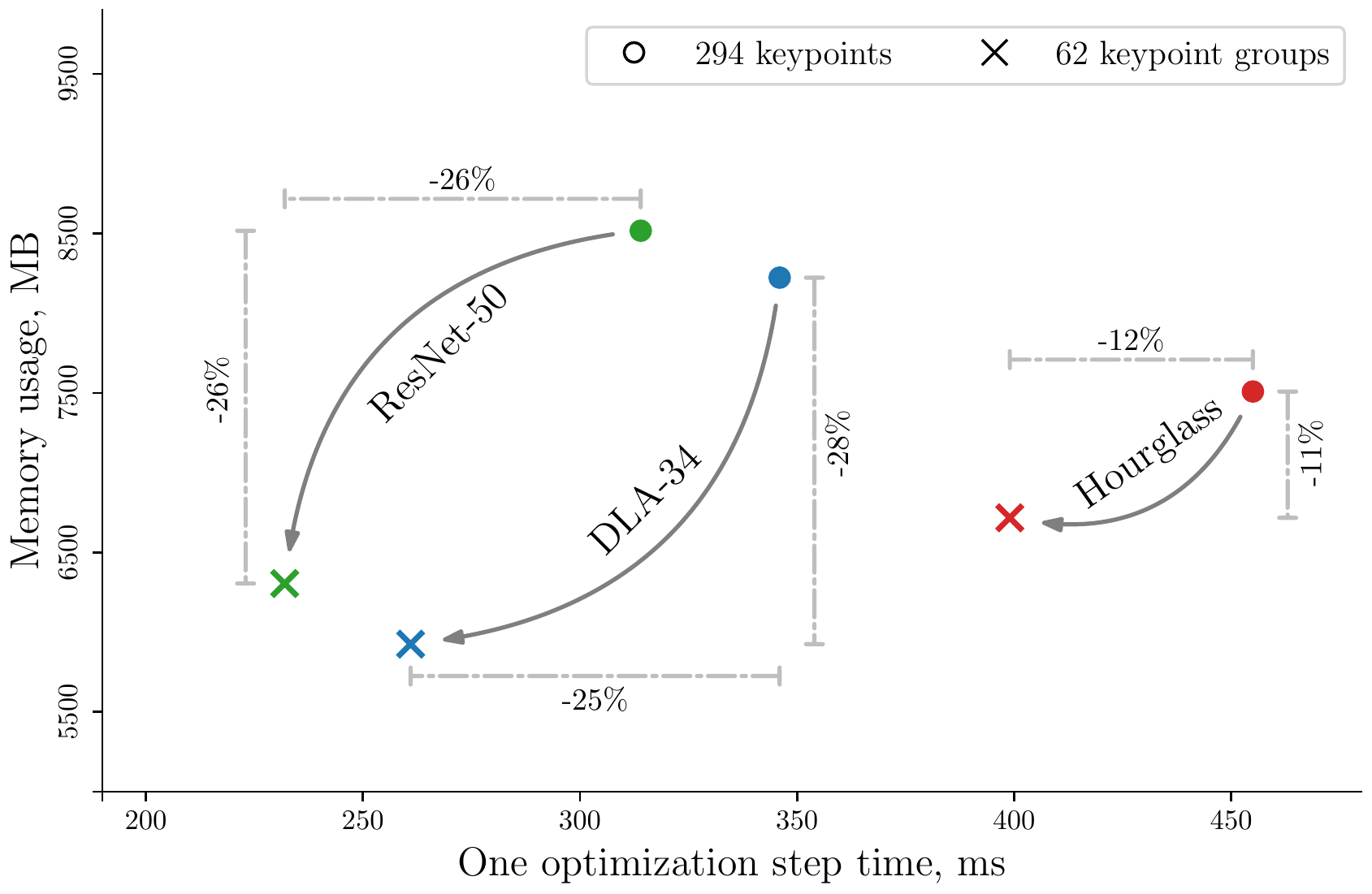}
	\caption{GPU memory consumption and training iteration time on RTX 2080ti.
		The input resolution is $\medmuskip=0mu 256\times256$, the batch size is 32 for both DLA-34 and ResNet-50, and 8 for Hourglass.
		Time in ms was measured for one optimization step: batch loading to GPU, forward pass, and backward pass.
		GPU memory was measured by using the nvidia-smi tool. The image is reproduced with permission from~\cite{DeepMarkCC}.}
	\label{fig:time-mem-plot}
\end{figure}

Memory consumption increases during neural network training (see Figure~\ref{fig:time-mem-plot}) because ground truth and loss computation are needed for every output channel.
In CenterNet implementation, focal loss for keypoint centers requires 18 extra operations and potential memory allocation, while regression loss for keypoint offsets requires only two extra operations.

\begin{table}
	\caption{Estimation of memory usage for different encoders and input resolution
	}
	\label{tab:activation_size}
	\begin{center}
		\begin{tabular}{|c|c|c|c|c|c|}
			\hline
			Encoder           & \makecell{Weights \\ (MB)} & \makecell{Activations \\ (MB)} & \makecell{Output\\tensors (MB)} & \makecell{Output\\tensors (\%)}  \\
			\hline
			\makecell{DLA-34\\$128\times128$}    & 74.4        & 17.0            & 3.5              & 3.8          \\
			\hline
			\makecell{DLA-34\\$256\times256$}    & 74.4        & 68.1            & 14.1              & 9.8          \\
			\hline
			\makecell{DLA-34\\$512\times512$}    & 74.4        & 272.6           & 56.3              & 16.1         \\
			\hline
			\makecell{ResNet-50\\$128\times128$} & 115.2       & 16.9            & 3.5              & 2.7          \\
			\hline
			\makecell{ResNet-50\\$256\times256$} & 115.2       & 67.8            & 14.1              & 7.7          \\
			\hline
			\makecell{ResNet-50\\$512\times512$} & 115.2       & 271.1           & 56.3              & 14.5         \\
			\hline
			\makecell{Hourglass\\$128\times128$} & 743.4       & 42.4            & 3.5                & 0.4          \\
			\hline
			\makecell{Hourglass\\$256\times256$} & 743.4       & 169.5           & 14.1               & 1.5          \\
			\hline
			\makecell{Hourglass\\$512\times512$} & 743.4       & 677.9           & 56.3               & 4.0          \\
			\hline
		\end{tabular}
	\end{center}
\end{table}

It is clear that certain clothing landmarks are a subset of others.
Therefore, for example, shorts do not require unique keypoints because they can be represented by a subset of trouser keypoints.
A manual grouping from~\cite{DeepMarkCC} allows 62 groups to be formed and reduces the number of output channels from 901 to $205 = 13 + 2 + 2 + 124 + 62 + 2$.
Figure~\ref{fig:time-mem-plot} illustrates the findings of the experiments, and reveals that such an approach can reduce memory consumption (during the training stage) and training time by up to 28\% and 26\%, respectively.
Theoretical evaluation (Table~\ref{tab:activation_size}) and experiment findings (Figure~\ref{fig:time-mem-plot}) reveal promising results for the keypoint grouping approach.

The key problem of manual grouping involves the human aspect.
To maintain accuracy, a smart, automatic approach is necessary.
An accurate grouping approach will benefit the generalization ability of the model as each group will receive more diverse training samples.
Meanwhile, the same grouping can be used as a tool for solving the class imbalance problem.

\subsection{Automatic Grouping Approach}

We view the keypoint grouping task as a clustering problem, which is defined as follows:
$k_i$ is a unique keypoint type where $i=\overline{1,n}$; and $g_j$ is a group label where $j=\overline{1,m}$.
For the DeepFashion2 dataset, $n$ is equal to 294; and for MS COCO Human Pose, $n$ is equal to 17.
The keypoint grouping task involves assigning a group label $g_j$ for every keypoint type $k_i$.

The first thing we must determine to solve the clustering problem is the dissimilarity measure between different keypoint types.
We propose a method that uses the following information about keypoints to measure distance:
\begin{itemize}
	\item Ground truth location of keypoints.
	\item Weights of the last convolution layer.
\end{itemize}

Ground truth location of keypoints can be directly used to analyze the spatial location of keypoints.
We evaluate each keypoint's offset from the center of an object, and then use this offset to estimate the mean location of every keypoint type.
Finally, we measure the dissimilarity between two keypoint types by calculating the Euclidean distance between mean locations.

Such an approach is very simple but does not utilize information about the content of the input image.
Moreover, this approach can only be used for solving keypoint regression task.

In fully convolution neural networks like CenterNet~\cite{CenterNet}, the last layer produces separate channels for each keypoint type.
For the regression task CenterNet provides each keypoint type with two channels ($\bigtriangleup$x and $\bigtriangleup$y), while the heatmap-based branch uses only one channel.
The last layer contains weights that are applied to the same activation feature map, and it produces results for different keypoint types.
We follow the intuition that similar keypoints should have convolution weights that are almost the same,
and use those weights to measure the Euclidean distance between different keypoint types.

The simplest way to implement keypoint grouping involves merging keypoints from different object classes into one group (for example, keypoints for shorts and trousers).
This can be achieved by setting a large distance value between keypoints from the same class, and will be referred to as grouping with restrictions.
Hierarchical clustering approaches can handle such keypoint merging well, and we use agglomerative clustering with linkage "average".

\begin{figure}[t]
	\centering
	\includegraphics[width=\columnwidth]{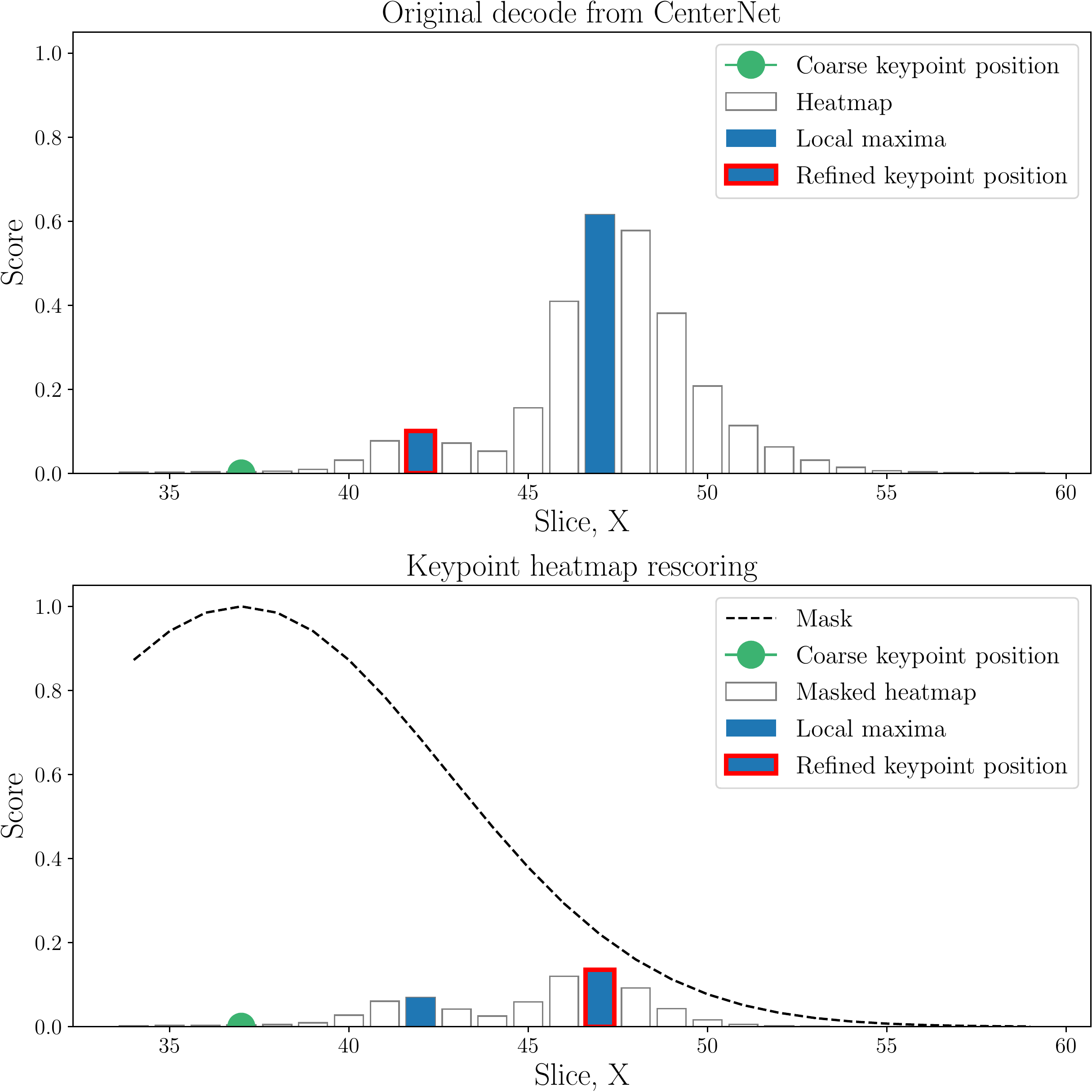}
	\caption{
	In CenterNet,
      the refined location is defined as the closest point to the coarse keypoint position, which in turn is defined as a local maximum on the heatmap.
     This can lead to errors when the closest prediction is not actually the best one (as evident in the diagram above).
     In the proposed technique, the refined location is determined as a global maximum of the keypoint heatmap rescored by the Gaussian mask, improving keypoint localization. Image is reproduced with permission from~\cite{DeepMarkCC}.
	}
	\label{fig:technique_4}
\end{figure}

To achieve the keypoint detection task, CenterNet relies on two branches: keypoint location regression to estimate approximate keypoint locations, and keypoint heatmap to refine regressed locations.
Given this, we can allow two keypoints from the same class (the same object) to be merged into a single cluster by regression, provided that they stay in different clusters by heatmap, and vice versa.
In this way, we can decode the original keypoints from clusters unambiguously.
This type of grouping method will be referred to as grouping without restrictions.

We discovered that decoding substantially affects accuracy, and provide the following modification to keypoint refinement from~\cite{DeepMarkCC}.

By rescoring the keypoint heatmap, we propose adding a penalty to the keypoint score in proportion to the distance from a coarse keypoint position with Gaussian function.
The final keypoint position is determined as a location of the maximum value in the rescored keypoint heatmap.

Assuming $mask$ is a heatmap with zero values by default,
we set $1$ into the $mask$ in the coarse keypoint position and fill neighbor values with 2D Gaussian function with standard deviation $sigma$ (see the second image in~Figure~\ref{fig:technique_4}).
The parameter $sigma$ is estimated using a subset of validation dataset for every model.

The keypoint heatmap is rescored for each coarse keypoint position through the following formula:
\begin{equation}
    \hat{H}_{kps} = H_{kps} \cdot mask \label{eq:4}
\end{equation}

This modified decoding is particularly important when we use grouping from the same class for keypoint regression.
The more points from the same class we merge, the further away the coarse keypoint position is from the correct point on the heatmap,
and the more likely the scenario in Figure~\ref{fig:technique_4} is to occur.
As we show in the next section, decoding with keypoint heatmap rescoring allows keypoints from the same class to be merged without significantly affecting accuracy.

\begin{figure}[t]
	\centering
	\includegraphics[width=\columnwidth]{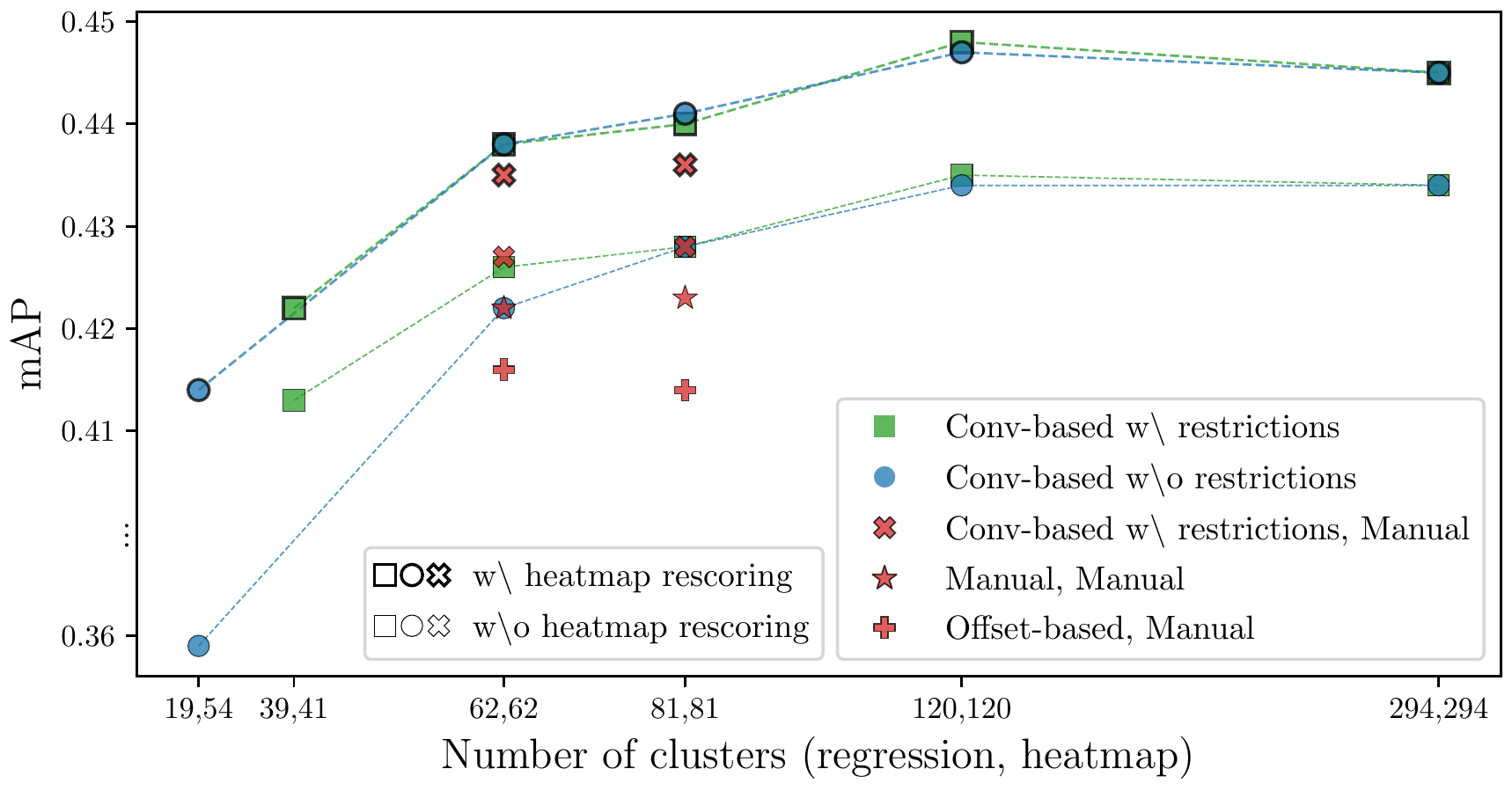}
	\caption{Comparison of grouping strategies on training from scratch.
		Manual groupings from~\cite{AggregationAF, DeepMarkCC} are used as-is and in combination
		with keypoint regression groupings: offsets-based and convolution-based (red points).
    Only the best results with heatmap rescoring for manual grouping are shown to avoid cluttering the chart. }
	\label{fig:groupings_comparison}
\end{figure}

\section{Experiments}

\begin{figure*}[t]
	\centering
	\includegraphics[width=\textwidth]{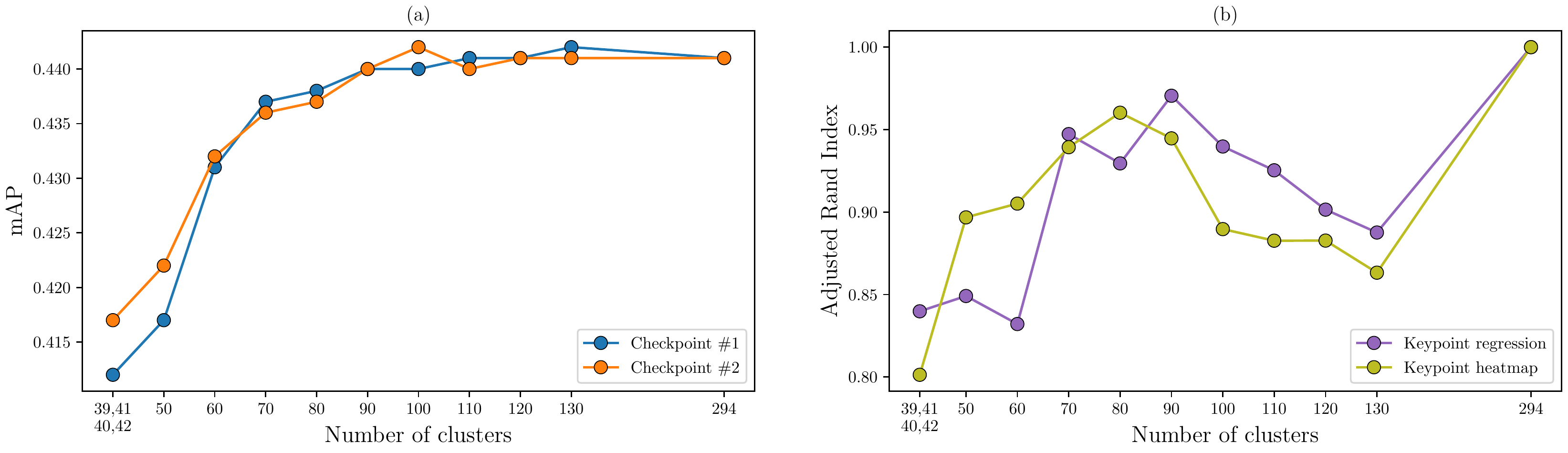}
	\caption{Grouping robustness.
		Figure (a) presents keypoint detection accuracy achieved after fine-tuning the initial model for 15 epochs with
		corresponding grouping obtained from two different checkpoints.
		Figure (b) illustrates a consensus between two groupings from different checkpoints: one for keypoint regression, and the other for keypoint heatmap.
		In almost all cases, the same number of clusters for regression and heatmap are used.
		The leftmost point presents a minimal number of clusters.}
	\label{fig:grouping_consistancy}
\end{figure*}

\subsection{Setup}

To demonstrate the effectiveness of the proposed approach and investigate its properties, we conducted experiments on the DeepFashion2 dataset~\cite{DeepFashion2}.
It provides annotations for 13 classes of clothing, with each class annotated by a unique set of keypoints (294 keypoints in total).
We also present the results for human pose estimation on the MS COCO dataset~\cite{lin2014microsoft}.

All experiments are performed with an input image resolution of $256\times256$, using a batch of 64 images and Adam optimizer.
Like the original CenterNet architecture, DLA-34~\cite{yu2018deep} is used as a backbone, but Deformable Convolution is replaced with conventional convolution layer for faster training.
Backbone weights are initialized by the model pre-trained on ImageNet.

An initial model was trained based on the following schedule.
The first 35 epochs with a constant learning rate of 3e-3 were trained,
and then a further 25 and 55 epochs for the DeepFashion2 dataset and the MS COCO dataset,
respectively, were trained, with the learning rate decaying exponentially to 1e-5.

Since detection quality varies slightly from epoch to epoch, the model was trained for another 20 short epochs of 250 iterations with a constant learning rate of 1e-5.
The best checkpoint on mini-val dataset was used for evaluation.

\subsection{Convolution-based grouping with restrictions}

\begin{table}
	\caption{DeepFashion2 keypoint detection accuracy with varying numbers of convolution channels before heads}
	\begin{center}
		\begin{tabular}{|c|c|c|c|c|}
			\hline
			Number of channels & 256 & 64 & 32 & 16 \\
			\hline
			mAP & 0.401	& 0.412 & \textbf{0.431} & 0.423 \\
			\hline
		\end{tabular}
	\end{center}
	\label{tab:num_channels}
\end{table}

In our experiments on the DeepFashion2 dataset, we use the modified loss function for learning the keypoint heatmap:
supervision is present only at the keypoint heatmap's channels corresponding to the points present in the image.
Therefore, we no longer penalize the network for non-zero values
in the channels corresponding to other classes, which in turn enables us to obtain closer weights for similar
points (collar of classes three and four, for example).
In this way, we obtain weights that reflect the semantic proximity of keypoints from different classes.

Furthermore, using the proposed loss function allows us to attain a more accurate model (with any grouping),
where the accuracy of the model with the original number of keypoints is 0.424 mAP and 0.434 mAP for the base and modified loss functions, respectively.
Meanwhile, the accuracy of the model with manual grouping by 62 groups is 0.417 mAP and 0.422 mAP for the base and modified loss functions, respectively.
The proposed loss function is only relevant for multi-class keypoint detection tasks such as DeepFashion2.
Experiments on the single class MS COCO Human Pose dataset did not yield any noticeable increase in accuracy.

Additionally, we discovered that the number of channels in the convolution layers
before heads significantly affects the quality of grouping approximation.
Originally, 256 channels were used to achieve higher detection accuracy, while
there were numerous channels.
This means, convolution weights may consist of noise,
and consequently affects keypoint approximation, ultimately leading to inefficient grouping.
Conversely, a significantly small number of channels cannot be adequate for providing accurate  keypoint representations.
In Table~\ref{tab:num_channels}, quantitative results are presented for grouping $(60,60)$ (the first value is a number of groups for keypoint regression,
and the second is a number of groups for keypoint heatmap) after fine-tuning for 15 epochs.
We found that 32 channels provide the most efficient grouping, and used this configuration for training.
Note that this configuration is only used to train a checkpoint to gather convolution weights for keypoints clustering.

To train a model with grouping we considered these two strategies:
\begin{itemize}
	\item Training from scratch with the parameters and schedule described in the beginning of this section;
	\item Fine-tuning from the original model for 15 epochs with a learning rate exponential decay from 4.0265e-4 to 1e-5.
	After keypoints are clustered, we cannot load head convolution weights directly.
      To deal with this issue we must initialize weights for each cluster $g$ as $W_{g} = \frac{1}{|g|} \sum_{k \in g} W_{k}$, where $k$ corresponds to the original keypoints.
\end{itemize}
Note that in both strategies, the model was trained with the original 256 channels to achieve a high detection score.

Furthermore, we investigated the robustness of convolution-based grouping in terms of detection accuracy and clustering consensus.
To achieve this, we evaluated various groupings from two independent checkpoints trained with the same parameters and schedule.
To evaluate accuracy, we fine-tuned models with each grouping for 15 epochs from the same checkpoint.
To measure the consensus between equivalent groupings, we used the Adjusted Rand Index~\cite{hubert1985comparing},
and the corresponding results are presented in Figure~\ref{fig:grouping_consistancy}.

\begin{figure}[t]
	\centering
	\includegraphics[width=\columnwidth]{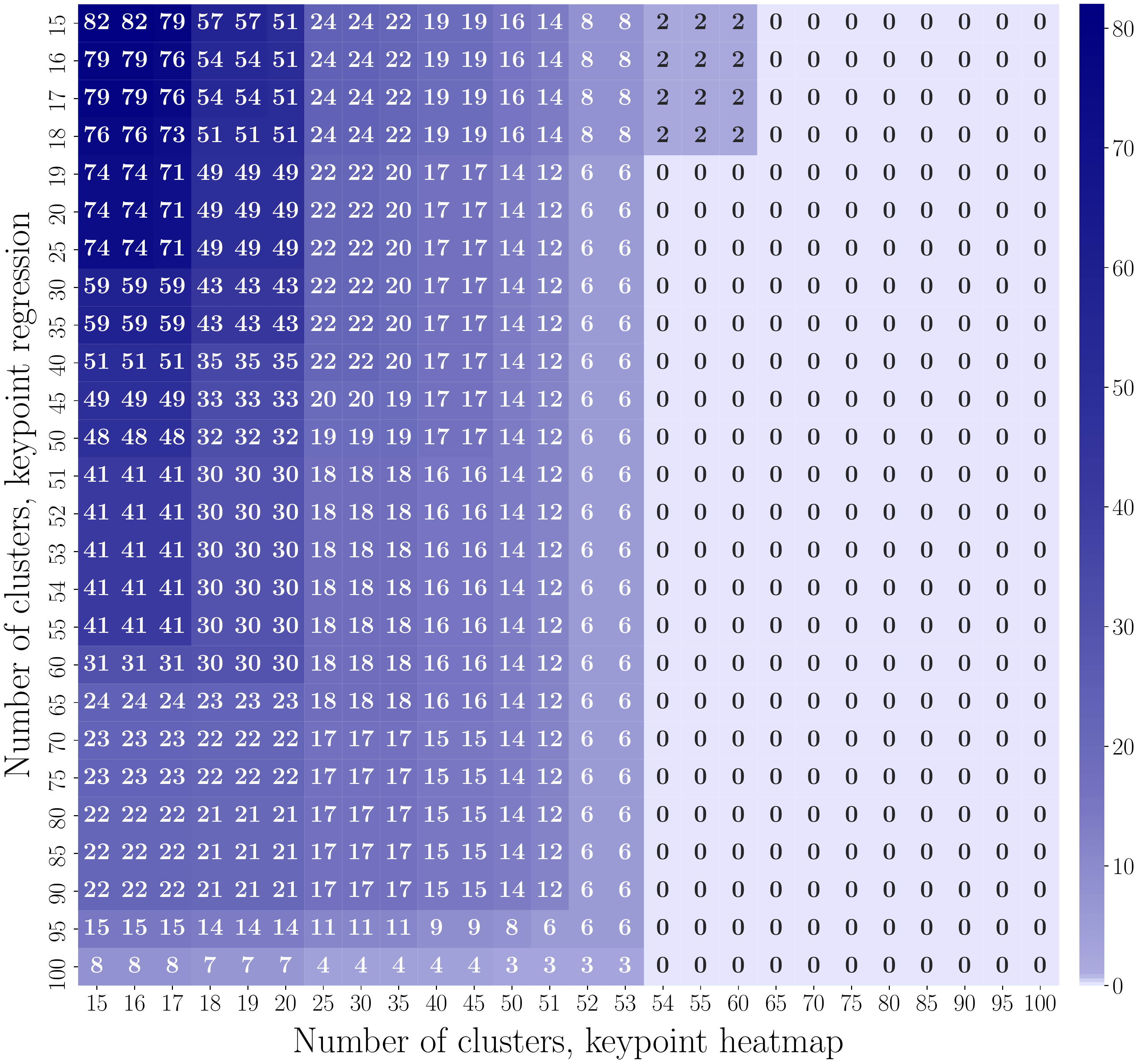}
	\caption{Matrix of ambiguous decoding.
		Contains the number of keypoint pairs from the same class that was merged for regression and heatmap simultaneously.}
	\label{fig:inconsistent_pairs_heatmap}
\end{figure}

While at a lower number of clusters, one can observe variations in accuracy and keypoint partitioning,
and groupings of 70 clusters become robust and able to achieve accuracy comparable to the original training.
A larger number causes the clusters to become very dense and small, and further partitioning is affected by variations in convolutional weights.
Therefore, the clustering consensus experiences a decrease, but this should not affect the effectiveness of grouping, which is supported by robust behavior in detection accuracy.
As such, the consensus measure can be used as a criterion to choose the number of clusters.

Finally, we trained the models with groupings from scratch and compared different grouping strategies, and the results are presented in Figure~\ref{fig:groupings_comparison}.
Grouping through the proposed approach is slightly more effective than manual grouping.
This improvement is driven mostly by the effective grouping for regression head since manual grouping has already been designed in a heatmap manner.
At the same time, straightforward clustering based on offsets provides less effective grouping which stresses the importance of keypoint approximation for clustering.
The heatmap rescoring technique increases accuracy in all cases.
This is particularly true when the high accuracy gain of 6 mAP is achieved for a very small number of clusters.
The best model with 120 clusters achieves slightly better accuracy compared to the naive approach.

We also trained lighter models -- DLA-34 with an input image resolution of 128$\times$128 and
MobileNet\_v2~\cite{sandler2018mobilenetv2} -- to gather weights and perform clustering.
We then used groupings $(62,62)$ and $(81,81)$ to train our standard model and achieved a detection accuracy very close to our predictions.
Moreover, we used groupings $(62,62)$ to train model DLA-34 with an input image resolution of $512\times512$ and achieved accuracy comparable to the original 294 keypoints model,
yet almost twice as fast while using 1.5 times less GPU memory for training.
These results demonstrate that the proposed grouping approach can be used on models that are small and geared towards training time to perform clustering.
The grouping can then be leveraged in the heavy model to accelerate the training process without compromising accuracy.

\subsection{Convolution-based grouping without restrictions}

The accuracy results  of the models that implement convolution-based grouping without restrictions on the keypoint detection task of the DeepFashion2 dataset are presented in Figure~\ref{fig:groupings_comparison}.
The permitted number of clusters for each output is determined by analyzing a matrix of ambiguous decoding
that contains the number of pairs from the same class that was merged for regression and heatmap simultaneously (Figure~\ref{fig:inconsistent_pairs_heatmap}).
The minimum grouping includes 19 and 54 clusters for the keypoint regression and keypoint heatmap, respectively.
The number of inconsistent pairs (pairs of keypoints corresponding to the same object category that were merged
into a single cluster) is shown in Table~\ref{tab:merged_pairs}.
As the number of clusters decreases, the number of inconsistent pairs for the keypoint regression increases,
which leads to a drop in accuracy relative to grouping with restrictions.
As shown in Figure~\ref{fig:groupings_comparison}, the keypoint heatmap rescoring technique prevents this drop from happening.

\begin{table}
	\caption{DeepFashion2: convolution-based grouping without restrictions and the number of inconsistent pairs}
	\label{tab:merged_pairs}
	\begin{center}
		\begin{tabular}{|c|c|c|c|c|}
			\hline
			 \makecell{Number of clusters}  &  \makecell{19,54} & \makecell{62,62} & \makecell{81,81} & \makecell{120,120} \\
			\hline
			\makecell{Keypoint regression} & 362  & 66 & 32 & 6 \\
                   \hline
			\makecell{Keypoint heatmap} & 13  & 5 & 2 & 0 \\
			\hline
		\end{tabular}
	\end{center}
\end{table}

Table~\ref{tab:exps_coco} shows the accuracy results for the keypont detection task on the MS COCO Human Pose dataset of the models
that implement convolution-based grouping without restrictions.
In these experiments, groupings were performed either for the keypoint regression or the keypoint heatmap.
Table~\ref{tab:exps_coco} also shows that the keypoint heatmap rescoring technique can significantly reduce the keypoint detection accuracy drop caused by a considerably smaller number of keypoint regression groups.
Our best grouping with 14 clusters achieves slightly better accuracy than the model with a full set of keypoints.

Let us analyze the results obtained for models with grouping for the keypoint heatmap.
If we merge the close points from the same class into one cluster, they will probably be in close proximity on the heatmap.
Close points on the same heatmap reduce accuracy;
therefore, if we change the grouping strategy and merge distant points in one group will accuracy increase?
To obtain the specified grouping we used clusterization by referring to the negative
distances between the average offsets of keypoints from the objects' centers
from annotations (agglomerative clustering with "complete" linkage was used), which will be referred to as "anti-offsets" grouping.
As shown in Table~\ref{tab:exps_coco}, this strategy allows us to obtain models with grouping for the keypoint heatmap featuring accuracy
close to that of the model with a full set of keypoints.

We applied 14 groups for keypoint regression to the human pose estimation model provided by the authors of CenterNet~\cite{CenterNet}.
We fine-tuned the original DLA-34 512$\times$512 model (with Deformable Convolution)
for 15 epochs using the batch of 16, with a learning rate exponential decay equal to 0.8 from 1.008125e-4.
The models have been tested with original image resolution and flip testing.
The accuracy of the model from the CenterNet paper is 0.589 AP with base decoding, and 0.596 AP with the keypoint heatmap rescoring technique;
whereas the accuracy of the obtained model is 0.579 AP with base decoding, and 0.592 AP with the keypoint heatmap rescoring technique.

To obtain convolutional weights for grouping, we trained the full network with six different outputs (heads) and losses.
However, this might not have been necessary and we can obtain good keypoint representation by training only one head -- regression or heatmap for corresponding grouping.
We trained two models with one head by utilizing the same parameters we used to train the full network.
With groupings obtained from those weights, we were able to achieve a detection accuracy that is lower by only 1.5 mAP compared to grouping from the full model.
Meanwhile, the chosen values of training parameters are not optimal for models with only one head, which resulted in over-fitting.
To improve the results, more appropriate parameter values can be used.

\begin{table}
      \caption{MS COCO Human Pose accuracy, AP}
	\label{tab:exps_coco}
	\setlength\tabcolsep{4.0pt}
	\begin{center}
		\begin{tabular}{|c|c|c|c|c|}
			\hline
			 \multicolumn{2}{|c|}{Number of clusters}  &  \makecell{10} & \makecell{14} & \makecell{17}  \\
			\hline
			 \multirow{2}*{\makecell{Keypoint regression,\\conv-based grouping}} &  \makecell{Base decode}
                                                                           & 0.322  & 0.349 & 0.359 \\
                                          \cline{2-2}
                                          & \makecell{Rescoring technique} & 0.360  & 0.369 &  0.368 \\
                     \hline
                     \multirow{2}*{\makecell{Keypoint heatmap,\\base decode}} &  \makecell{Conv-based grouping}
                                                                             & 0.281  & 0.334 & 0.359 \\
                                          \cline{2-2}
                                          & \makecell{Anti-offsets grouping} & 0.356  & 0.356 &  0.359 \\
                     \hline
		\end{tabular}
	\end{center}
\end{table}

\section{Conclusion}

We have shown that the proposed automatic grouping approach with the special post-processing technique
works with the CenterNet architecture on human pose
estimation and fashion landmark detection tasks.
It boosts training speed, accelerates inference time, and reduces memory consumption without compromising accuracy.

The proposed grouping approach can be used on models that are small and geared towards training time to perform clustering,
and then the grouping can be used in the heavy model to accelerate the training process.

Heatmap is the standard coordinate representation in keypoint detection tasks~\cite{DARK}.
For this reason, the proposed approach can also be applied to almost any keypoint detection architecture,
for example, HRNet~\cite{wang2020deep}, PoseFix~\cite{moon2018posefix}, and Mask R-CNN~\cite{MaskRCNN}.

\bibliographystyle{IEEEtran}
\bibliography{IEEEabrv,bibliography}

\begin{thebibliography}{10}
\providecommand{\url}[1]{#1}
\csname url@samestyle\endcsname
\providecommand{\newblock}{\relax}
\providecommand{\bibinfo}[2]{#2}
\providecommand{\BIBentrySTDinterwordspacing}{\spaceskip=0pt\relax}
\providecommand{\BIBentryALTinterwordstretchfactor}{4}
\providecommand{\BIBentryALTinterwordspacing}{\spaceskip=\fontdimen2\font plus
\BIBentryALTinterwordstretchfactor\fontdimen3\font minus
  \fontdimen4\font\relax}
\providecommand{\BIBforeignlanguage}[2]{{%
\expandafter\ifx\csname l@#1\endcsname\relax
\typeout{** WARNING: IEEEtran.bst: No hyphenation pattern has been}%
\typeout{** loaded for the language `#1'. Using the pattern for}%
\typeout{** the default language instead.}%
\else
\language=\csname l@#1\endcsname
\fi
#2}}
\providecommand{\BIBdecl}{\relax}
\BIBdecl

\bibitem{DeepFashion2}
Y.~Ge, R.~Zhang, X.~Wang, X.~Tang, and P.~Luo, ``{DeepFashion2}: A versatile
  benchmark for detection, pose estimation, segmentation and re-identification
  of clothing images,'' in \emph{Proceedings of the IEEE Conference on Computer
  Vision and Pattern Recognition}, 2019, pp. 5337--5345.

\bibitem{CenterNet}
\BIBentryALTinterwordspacing
X.~Zhou, D.~Wang, and P.~Kr{\"{a}}henb{\"{u}}hl, ``Objects as points,''
  \emph{CoRR}, vol. abs/1904.07850, 2019. [Online]. Available:
  \url{http://arxiv.org/abs/1904.07850}
\BIBentrySTDinterwordspacing

\bibitem{DeepMarkCC}
\BIBentryALTinterwordspacing
A.~Sidnev, A.~Krapivin, A.~Trushkov, E.~Krasikova, M.~Kazakov, and M.~Viryasov,
  ``Deepmark++: Real-time clothing detection at the edge,'' \emph{CoRR}, vol.
  abs/2006.00710, 2020. [Online]. Available:
  \url{https://arxiv.org/abs/2006.00710}
\BIBentrySTDinterwordspacing

\bibitem{alibaba2020mnn}
X.~Jiang, H.~Wang, Y.~Chen, Z.~Wu, L.~Wang, B.~Zou, Y.~Yang, Z.~Cui, Y.~Cai,
  T.~Yu, C.~Lv, and Z.~Wu, ``Mnn: A universal and efficient inference engine,''
  in \emph{MLSys}, 2020.

\bibitem{AggregationAF}
\BIBentryALTinterwordspacing
T.~Lin, ``Aggregation and finetuning for clothes landmark detection,''
  \emph{CoRR}, vol. abs/2005.00419, 2020. [Online]. Available:
  \url{https://arxiv.org/abs/2005.00419}
\BIBentrySTDinterwordspacing

\bibitem{lin2014microsoft}
T.-Y. Lin, M.~Maire, S.~Belongie, J.~Hays, P.~Perona, D.~Ramanan,
  P.~Doll{\'a}r, and C.~L. Zitnick, ``Microsoft coco: Common objects in
  context,'' in \emph{European conference on computer vision}.\hskip 1em plus
  0.5em minus 0.4em\relax Springer, 2014, pp. 740--755.

\bibitem{alej2016stacked}
A.~Newell, K.~Yang, and J.~Deng, ``Stacked hourglass networks for human pose
  estimation,'' 2016.

\bibitem{zhang2014facial}
Z.~Zhang, P.~Luo, C.~C. Loy, and X.~Tang, ``Facial landmark detection by deep
  multi-task learning,'' in \emph{European conference on computer
  vision}.\hskip 1em plus 0.5em minus 0.4em\relax Springer, 2014, pp. 94--108.

\bibitem{liu2016deepfashion}
Z.~Liu, P.~Luo, S.~Qiu, X.~Wang, and X.~Tang, ``Deepfashion: Powering robust
  clothes recognition and retrieval with rich annotations,'' in
  \emph{Proceedings of the IEEE conference on computer vision and pattern
  recognition}, 2016, pp. 1096--1104.

\bibitem{zheng2018modanet}
S.~Zheng, F.~Yang, M.~H. Kiapour, and R.~Piramuthu, ``Modanet: A large-scale
  street fashion dataset with polygon annotations,'' in \emph{2018 ACM
  Multimedia Conference on Multimedia Conference}.\hskip 1em plus 0.5em minus
  0.4em\relax ACM, 2018, pp. 1670--1678.

\bibitem{fang2017rmpe}
H.-S. Fang, S.~Xie, Y.-W. Tai, and C.~Lu, ``Rmpe: Regional multi-person pose
  estimation,'' in \emph{Proceedings of the IEEE International Conference on
  Computer Vision}, 2017, pp. 2334--2343.

\bibitem{MaskRCNN}
\BIBentryALTinterwordspacing
K.~He, G.~Gkioxari, P.~Doll{\'{a}}r, and R.~B. Girshick, ``Mask {R-CNN},''
  \emph{CoRR}, vol. abs/1703.06870, 2017. [Online]. Available:
  \url{http://arxiv.org/abs/1703.06870}
\BIBentrySTDinterwordspacing

\bibitem{cao2018openpose}
Z.~Cao, G.~Hidalgo, T.~Simon, S.-E. Wei, and Y.~Sheikh, ``Openpose: realtime
  multi-person 2d pose estimation using part affinity fields,'' \emph{arXiv
  preprint arXiv:1812.08008}, 2018.

\bibitem{pishchulin2016deepcut}
L.~Pishchulin, E.~Insafutdinov, S.~Tang, B.~Andres, M.~Andriluka, P.~V. Gehler,
  and B.~Schiele, ``Deepcut: Joint subset partition and labeling for multi
  person pose estimation,'' in \emph{Proceedings of the IEEE conference on
  computer vision and pattern recognition}, 2016, pp. 4929--4937.

\bibitem{liu2016ssd}
W.~Liu, D.~Anguelov, D.~Erhan, C.~Szegedy, S.~Reed, C.-Y. Fu, and A.~C. Berg,
  ``Ssd: Single shot multibox detector,'' in \emph{European conference on
  computer vision}.\hskip 1em plus 0.5em minus 0.4em\relax Springer, 2016, pp.
  21--37.

\bibitem{redmon2016you}
J.~Redmon, S.~Divvala, R.~Girshick, and A.~Farhadi, ``You only look once:
  Unified, real-time object detection,'' in \emph{Proceedings of the IEEE
  conference on computer vision and pattern recognition}, 2016, pp. 779--788.

\bibitem{law2018cornernet}
H.~Law and J.~Deng, ``Cornernet: Detecting objects as paired keypoints,'' in
  \emph{Proceedings of the European Conference on Computer Vision (ECCV)},
  2018, pp. 734--750.

\bibitem{marszalek2007semantic}
M.~Marszalek and C.~Schmid, ``Semantic hierarchies for visual object
  recognition,'' in \emph{2007 IEEE Conference on Computer Vision and Pattern
  Recognition}.\hskip 1em plus 0.5em minus 0.4em\relax IEEE, 2007, pp. 1--7.

\bibitem{deng2014large}
J.~Deng, N.~Ding, Y.~Jia, A.~Frome, K.~Murphy, S.~Bengio, Y.~Li, H.~Neven, and
  H.~Adam, ``Large-scale object classification using label relation graphs,''
  in \emph{European conference on computer vision}.\hskip 1em plus 0.5em minus
  0.4em\relax Springer, 2014, pp. 48--64.

\bibitem{zhang2016embedding}
X.~Zhang, F.~Zhou, Y.~Lin, and S.~Zhang, ``Embedding label structures for
  fine-grained feature representation,'' in \emph{Proceedings of the IEEE
  Conference on Computer Vision and Pattern Recognition}, 2016, pp. 1114--1123.

\bibitem{verma2012learning}
N.~Verma, D.~Mahajan, S.~Sellamanickam, and V.~Nair, ``Learning hierarchical
  similarity metrics,'' in \emph{2012 IEEE conference on computer vision and
  pattern recognition}.\hskip 1em plus 0.5em minus 0.4em\relax IEEE, 2012, pp.
  2280--2287.

\bibitem{deng2010does}
J.~Deng, A.~C. Berg, K.~Li, and L.~Fei-Fei, ``What does classifying more than
  10,000 image categories tell us?'' in \emph{European conference on computer
  vision}.\hskip 1em plus 0.5em minus 0.4em\relax Springer, 2010, pp. 71--84.

\bibitem{zhao2011large}
B.~Zhao, F.~Li, and E.~P. Xing, ``Large-scale category structure aware image
  categorization,'' in \emph{Advances in Neural Information Processing
  Systems}, 2011, pp. 1251--1259.

\bibitem{yu2018deep}
F.~Yu, D.~Wang, E.~Shelhamer, and T.~Darrell, ``Deep layer aggregation,'' in
  \emph{Proceedings of the IEEE conference on computer vision and pattern
  recognition}, 2018, pp. 2403--2412.

\bibitem{hubert1985comparing}
L.~Hubert and P.~Arabie, ``Comparing partitions,'' \emph{Journal of
  classification}, vol.~2, no.~1, pp. 193--218, 1985.

\bibitem{sandler2018mobilenetv2}
M.~Sandler, A.~Howard, M.~Zhu, A.~Zhmoginov, and L.-C. Chen, ``Mobilenetv2:
  Inverted residuals and linear bottlenecks,'' in \emph{Proceedings of the IEEE
  conference on computer vision and pattern recognition}, 2018, pp. 4510--4520.

\bibitem{DARK}
F.~Zhang, X.~Zhu, H.~Dai, M.~Ye, and C.~Zhu, ``Distribution-aware coordinate
  representation for human pose estimation,'' \emph{arXiv preprint
  arXiv:1910.06278}, 2019.

\bibitem{wang2020deep}
J.~Wang, K.~Sun, T.~Cheng, B.~Jiang, C.~Deng, Y.~Zhao, D.~Liu, Y.~Mu, M.~Tan,
  X.~Wang \emph{et~al.}, ``Deep high-resolution representation learning for
  visual recognition,'' \emph{IEEE transactions on pattern analysis and machine
  intelligence}, 2020.

\bibitem{moon2018posefix}
G.~Moon, J.~Y. Chang, and K.~M. Lee, ``Posefix: Model-agnostic general human
  pose refinement network,'' 2018.

\end{thebibliography}

\end{document}